\documentclass[runningheads]{llncs}

\usepackage{eccv}

\usepackage{eccvabbrv}

\usepackage{graphicx}
\usepackage{booktabs}
\usepackage{makecell}
\usepackage{dsfont}
\usepackage[accsupp]{axessibility}  

\usepackage{hyperref}

\usepackage{orcidlink}

\begin{document}

\title{ReinDriveGen: Reinforcement Post-Training for Out-of-Distribution Driving Scene Generation} 

\titlerunning{Abbreviated paper title}

\author{
    Hao Zhang$^{1}$, \quad Lue Fan$^{1,2}$, \quad Weikang Bian$^{1}$, \\
    Zehuan Wu$^{3}$, \quad Lewei Lu$^{3}$, \quad Zhaoxiang Zhang$^{2}$, \quad Hongsheng Li$^{1}$ \\[0.5em]
}

 
\institute{ {\small $^{1}$MMLab, CUHK \qquad $^{2}$CASIA} \qquad
    {\small  $^{3}$SenseTime Research} \\
    {\small \textbf{Project page:} \textcolor{blue}{https://drive-sim.github.io/ReinDriveGen/}}
    }

\maketitle

\begin{figure}[htbp]
  \centering
  \includegraphics[width=\linewidth]{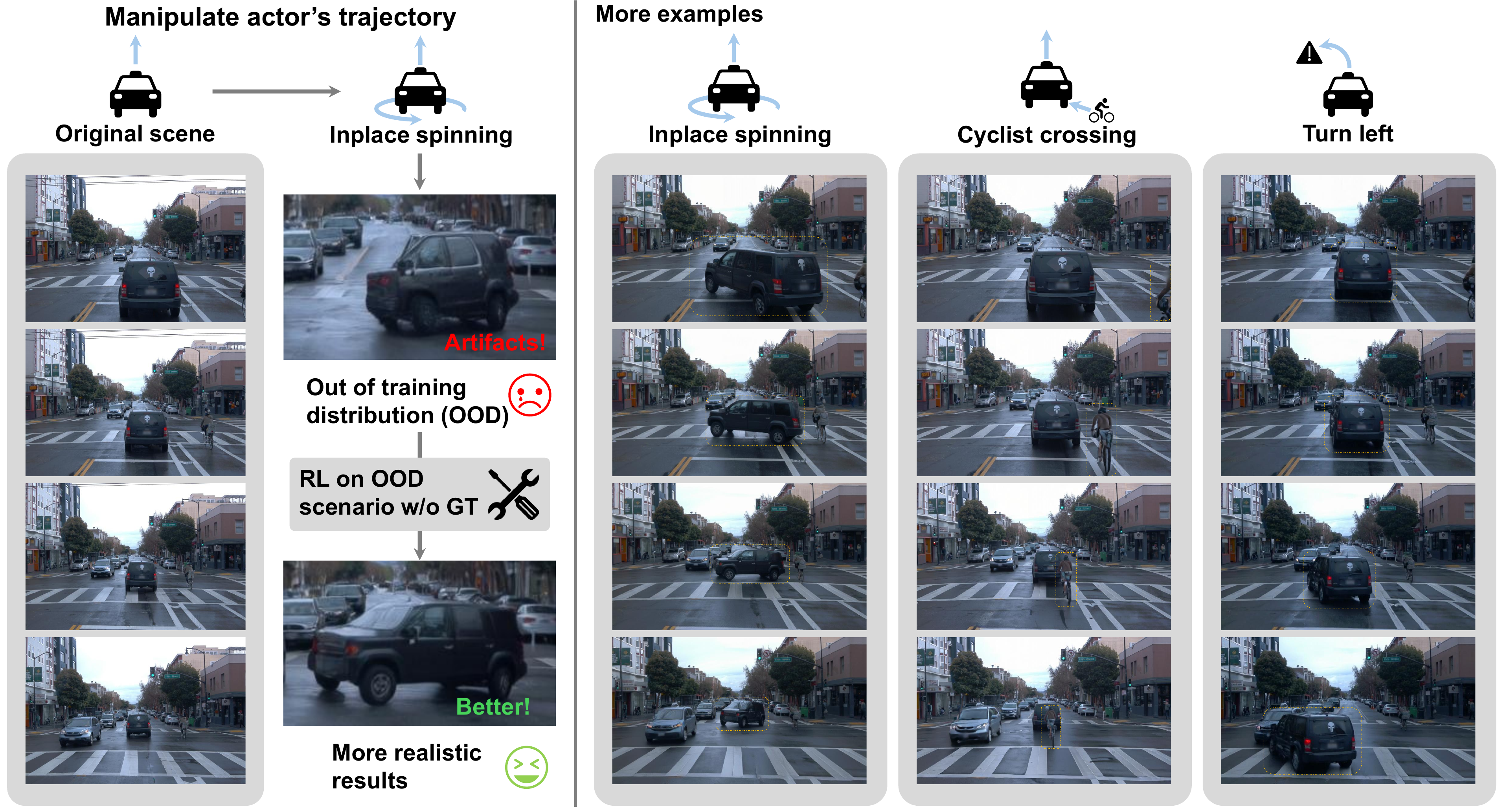}

  \caption{\textbf{ReinDriveGen enables photorealistic generation of OOD driving scenarios.} ReinDriveGen can manipulate actor trajectories to synthesize diverse safety-critical corner cases such as vehicle in-place spinning, cyclist crossing, and left-turn collisions. Our RL-based post-training significantly improves generation quality for these OOD edits without requiring ground-truth supervision.}
  \vspace{-0.3in}

  \label{fig:teaser}
\end{figure}

\begin{abstract}

We present ReinDriveGen, a framework that enables full controllability over dynamic driving scenes, allowing users to freely edit actor trajectories to simulate safety-critical corner cases such as front-vehicle collisions, drifting cars, vehicles spinning out of control, pedestrians jaywalking, and cyclists cutting across lanes. Our approach constructs a dynamic 3D point cloud scene from multi-frame LiDAR data, introduces a vehicle completion module to reconstruct full 360° geometry from partial observations, and renders the edited scene into 2D condition images that guide a video diffusion model to synthesize realistic driving videos. Since such edited scenarios inevitably fall outside the training distribution, we further propose an RL-based post-training strategy with a pairwise preference model and a pairwise reward mechanism, enabling robust quality improvement under out-of-distribution conditions without ground-truth supervision. Extensive experiments demonstrate that ReinDriveGen outperforms existing approaches on edited driving scenarios and achieves state-of-the-art results on novel ego viewpoint synthesis.

\end{abstract}

\section{Introduction}
\label{sec:intro}
Photorealistic driving scene simulation is indispensable for the development and validation of autonomous driving systems. A high-fidelity simulator allows arbitrary editing of driving scenes to generate diverse and realistic scenarios, including rare but safety-critical corner cases such as sudden cut-ins, near-collisions, vehicle drifting, and pedestrians jaywalking, thereby facilitating efficient data augmentation and thorough evaluation of driving policies.


A central challenge in building such a simulator lies in the inevitable \textbf{distribution gap} between training and inference. Current approaches, whether video generation-based~\cite{gao2025magicdrive-v2,ni2025maskgwm,nvidia2026worldsimulationvideofoundation} or reconstruction-based~\cite{mildenhall2020nerf,kerbl20233d,zhao2024drive,fan2025freesim,ni2025recondreamer,yan2025streetcrafter}, are fundamentally trained or optimized on recorded driving data. At inference time, however, a useful simulator must support arbitrary edits, such as repositioning vehicles, altering trajectories, and rendering from novel ego viewpoints, all of which produce conditions that deviate significantly from the training distribution. This gap is further exacerbated for safety-critical corner cases such as vehicles spinning in place, rollovers, and head-on collisions, which are inherently rare or entirely absent in normal driving logs, leaving the model with few or no corresponding training examples. Standard supervised fine-tuning (SFT) cannot close this gap, because ground-truth video only exists for the original recorded trajectories and viewpoints; no real reference can be obtained for any edited configuration. This out-of training distribution (OOD) degradation is most pronounced on vehicles, as they are the primary objects being edited and thus undergo the most drastic distribution shift, making vehicle quality the primary bottleneck for realistic simulation of safety-critical corner cases.


To address this fundamental challenge, we propose ReinDriveGen, a unified framework whose core insight is to leverage \textbf{reinforcement learning (RL) based post-training} to improve generation quality under OOD conditions, going beyond what SFT alone can achieve. Our framework comprises two components. The first is a Point Cloud Conditioned Video Diffusion Simulator. We aggregate and colorize multi-frame LiDAR point clouds to build a dynamic 3D scene, and introduce a vehicle completion module to reconstruct the full 360° geometry of partially observed vehicles. The completed scene is then rendered into 2D pseudo-images that serve as conditions for a video diffusion model to synthesize photorealistic driving videos.

The second is RL-based post-training, which finetunes the video diffusion model to bridge the distribution gap that SFT cannot close, enabling high-quality generation under OOD conditions. We extend DiffusionNFT~\cite{zheng2025diffusionnft} from image generation to video generation as our RL post-training framework. For each OOD condition, the framework generates a group of candidate videos and uses a reward model to partition them into positive and negative subsets, defining a contrastive improvement direction via flow matching~\cite{lipman2022flow}. A critical question is how to design the reward model, as no existing reward model is tailored to evaluate the quality of generated driving vehicles. A straightforward approach is to train a pointwise reward model that assigns an absolute scalar score to each sample under the Bradley-Terry preference framework. However, such pointwise scoring is susceptible to reward hacking, where negligible score differences are disproportionately amplified into misleading training signals~\cite{Pref-GRPO&UniGenBench}. Moreover, pairwise comparison provides more robust judgments than evaluating each sample in isolation, as the model can directly contrast two samples rather than relying on absolute scores prone to systematic biases. Building on this insight, we curate a dataset of positive and negative driving vehicle pairs and train a dedicated \textbf{pairwise preference model}. To convert pairwise outcomes into per-sample rewards, we propose a \textbf{pairwise reward mechanism}: for each group of candidates, we perform all pairwise comparisons, and a sample's reward is determined by how many comparisons it wins, with frequent winners receiving high rewards and frequent losers being penalized accordingly. This relative evaluation scheme avoids the pitfalls of absolute scoring and yields robust reward signals. Extensive experiments demonstrate that our RL-based post-training brings substantial quality improvements on OOD edited scenarios.

Our main contributions are summarized as follows: (1) We propose a Point Cloud-Conditioned Video Diffusion Simulator capable of simulating edited actor trajectories and novel ego-vehicle viewpoints. (2) We extend RL-based post-training from image to video generation and propose a pairwise reward mechanism with a dedicated pairwise preference model, yielding robust training signals that effectively address vehicle quality degradation under out-of-distribution driving scenarios.

\section{Related Work}

\textbf{Reconstruction-Based Driving Simulators.}
Neural Radiance Fields~\cite{mildenhall2020nerf} and 3D Gaussian Splatting~\cite{kerbl20233d} have driven rapid progress in 3D scene reconstruction, and numerous works have adapted these representations to dynamic driving scenarios~\cite{huang2024s3gaussian,chen2026periodic,wu2023mars,guo2023streetsurf,turki2023suds,yang2023unisim,tonderski2024neurad,zhou2024drivinggaussian,yan2024street}. However, these reconstruction-centric methods are intrinsically limited to the training trajectory and struggle to extrapolate high-fidelity views when the camera deviates significantly from the recorded path. Recent hybrid approaches~\cite{fan2025freesim,zhao2024drive,ni2025recondreamer,yan2025streetcrafter} integrate video diffusion priors to improve novel-viewpoint rendering, yet balancing generation and reconstruction remains non-trivial, often leading to geometric artifacts or blurred details. More fundamentally, all reconstruction-based methods cannot properly handle vehicle trajectory editing: repositioning a vehicle exposes previously unobserved regions that produce visible holes, while baked-in visual effects such as lighting and shadows cannot be re-synthesized for the edited configuration.

\noindent \textbf{Diffusion-Based Driving Simulators.}
In parallel with reconstruction-based methods, diffusion models have emerged as powerful tools for driving simulation. Approaches such as MagicDrive-V2~\cite{gao2025magicdrive-v2}, Vista~\cite{gao2024vista}, MaskGWM~\cite{ni2025maskgwm}, and Cosmos~\cite{nvidia2026worldsimulationvideofoundation} synthesize driving videos from structured inputs such as text prompts, HD maps, and bounding boxes. However, these video-generation models lack an explicit underlying 3D representation, which precludes precise control over camera trajectories and makes it difficult to maintain consistent 3D scene structure across frames. Furthermore, due to the inherent stochasticity of the diffusion sampling process, the generated content often varies significantly across different runs, making it difficult to perform multiple rounds of distinct edits—such as modifying different vehicles' trajectories—while keeping the rest of the scene fixed and consistent. More fundamentally, constrained by their training data distribution, these methods struggle to faithfully simulate unusual driving behaviors such as vehicle drifting or in-place spins that are rarely represented in standard driving datasets. Although FreeVS~\cite{wang2024freevs} attempts to mitigate control issues by leveraging LiDAR point clouds for geometric guidance, it remains limited by the sensor's range and fails to recover visual content outside the LiDAR scan, leading to incomplete scene synthesis.

\begin{figure*}[t]
    \centering
    \includegraphics[width=\linewidth]{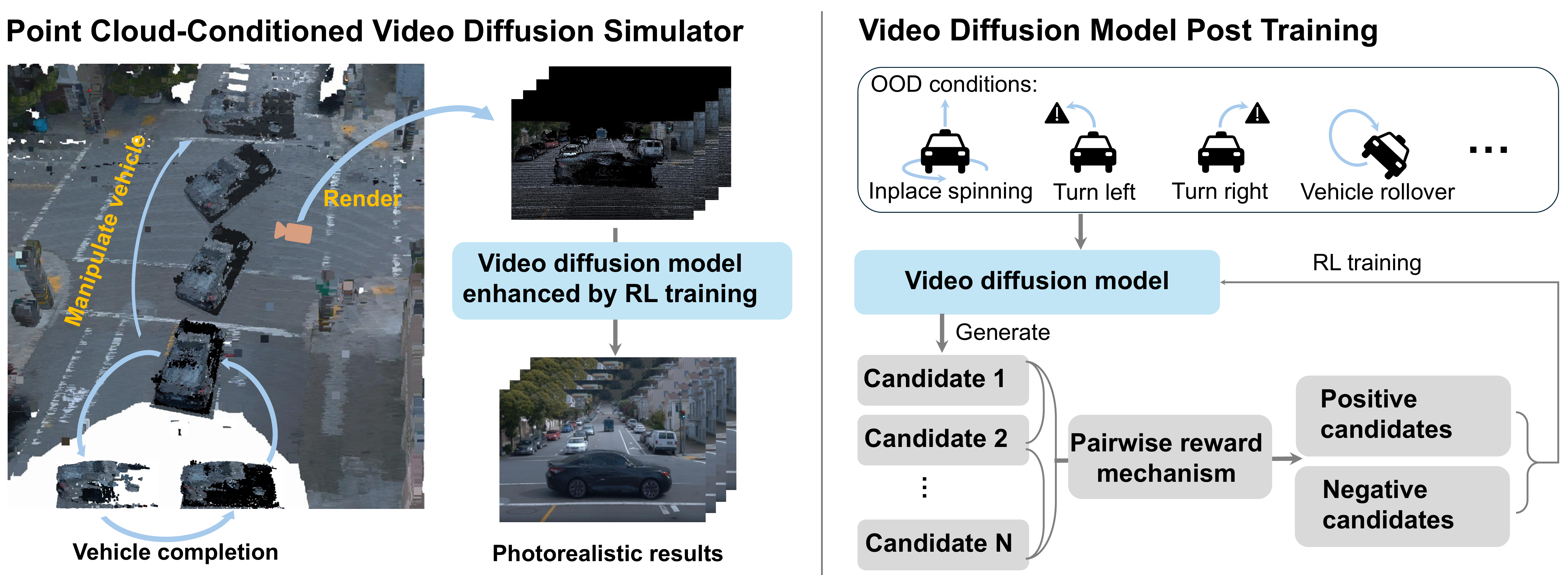}
    \vspace{-0.2in}
    \caption{\textbf{Overview of ReinDriveGen's two core components.} \textbf{Left:} Our simulator edits dynamic point clouds and completes vehicle geometry to render structural pseudo-images, which condition the video diffusion model for photorealistic synthesis. \textbf{Right:} To enhance quality in out-of-distribution (OOD) scenarios, we employ RL post-training. A pairwise reward mechanism ranks generated candidates into positive and negative sets, providing robust contrastive supervision.}
    \label{fig:main_pipeline}
\end{figure*}

\section{Method}
Our method consists of two core components as illustrated in Fig. \ref{fig:main_pipeline}: a Point Cloud-Conditioned Video Diffusion Simulator (Sec. \ref{sec:pc_sim}) and an RL-based post-training framework (Sec.~\ref{sec:rl}).

\subsection{Point Cloud-Conditioned Video Diffusion Simulator}
\label{sec:pc_sim}

Without an explicit 3D scene representation, video diffusion models often struggle to maintain multi-view geometric consistency, and it remains challenging to perform multiple rounds of targeted scene edits while keeping the rest of the environment intact. We therefore adopt dynamic 3D point clouds as our 3D backbone and render them into 2D pseudo-images that provide pixel-level conditions for a video diffusion model, inspired by ~\cite{wang2024freevs,yan2025streetcrafter}.

\subsubsection{Dynamic 3D Point Cloud Construction.}
A single LiDAR scan is too sparse to serve as a reliable condition. We therefore aggregate points across frames in a category-aware manner, leveraging the annotated 3D bounding boxes provided by Waymo~\cite{Sun_2020_CVPR} for all dynamic actors (vehicles, pedestrians, and cyclists). For the static background, we accumulate all 200 frames of each scene into a unified world coordinate system. For each vehicle, we transform its per-frame LiDAR points into the object's canonical frame using the annotated bounding boxes and aggregate all frames. Pedestrians and cyclists are deformable, so we only aggregate $\pm$2 frames (5 in total) around each time step. Each point is colorized by projecting it onto the corresponding camera image and retrieving the RGB value.

\subsubsection{Vehicle Completion.}
Even after multi-frame aggregation, vehicles are often only partially observed due to occlusion and limited recorded viewpoints. 
We apply AdaPoinTr~\cite{yu2021pointr}, a point cloud completion method, to reconstruct full 360° vehicle geometry. The model is initialized from a ShapeNet-55 pre-trained checkpoint and fine-tuned on Projected ShapeNet-55 (car category only; see supplementary for details). After completion, each vehicle can be freely repositioned and observed from arbitrary viewpoints without geometric holes, ensuring multi-view geometric consistency across edits. Note that the completed points carry no color information and are assigned black; we leave the texture synthesis of these surfaces to the downstream video diffusion model.

\subsubsection{Scene Editing and Rendering.}
\label{sec:edit_render}
With the dynamic 3D point cloud augmented by completed vehicles in place, we can freely edit actor trajectories and ego-vehicle viewpoints within the reconstructed scene. The edited point cloud is then rendered into 2D pseudo-images via projection; regions not covered by any LiDAR point are left black.

\subsubsection{Conditional Video Generation.}
\label{sec:cond_gen}

The pseudo-images rendered from the edited point cloud provide geometric guidance but are inevitably noisy and incomplete due to LiDAR sparsity, 
multi-frame accumulation artifacts, and untextured completed surfaces. A video diffusion model is employed to transform these coarse renderings 
into photorealistic driving videos.

To ensure that the regions beyond LiDAR coverage remain visually consistent across different edits, we anchor each generated sequence to a recorded frame from the original capture, which serves as a reference image providing appearance priors. Conditioned on both the reference image and the rendered pseudo-image sequence, the video diffusion model fills in LiDAR-unobserved regions guided by the reference image, corrects multi-frame accumulation artifacts, synthesizes 
realistic textures on the completed but untextured vehicle surfaces, and generates plausible lighting and shadow effects.

We adopt the VACE-1.3B~\cite{jiang2025vace} model as our video diffusion backbone. The reference image and pseudo-image sequence are injected through VACE's Video Condition Unit (VCU) as a composition of Reference-to-Video Generation (R2V) and Video-to-Video Editing (V2V). We jointly fine-tune the Context Adapter and the DiT on the Waymo dataset using paired pseudo-image sequences and recorded ground-truth videos. At inference time, novel ego-vehicle viewpoints often expose completed but untextured vehicle surfaces that never appear under the original recorded views. To improve robustness to such cases, we adopt the idea of pseudo-view simulation from GA-Drive~\cite{zhang2026ga}: we randomly mask regions adjacent to depth discontinuities, simulating the black untextured areas that emerge under novel viewpoints. We further apply random rectangular block masking to handle arbitrary occlusion patterns. Training videos are 49 frames at 480$\times$832 resolution, and the text prompt is fixed as ``A realistic autonomous driving scene.''

\begin{figure*}[t]
    \centering
    \includegraphics[width=\linewidth]{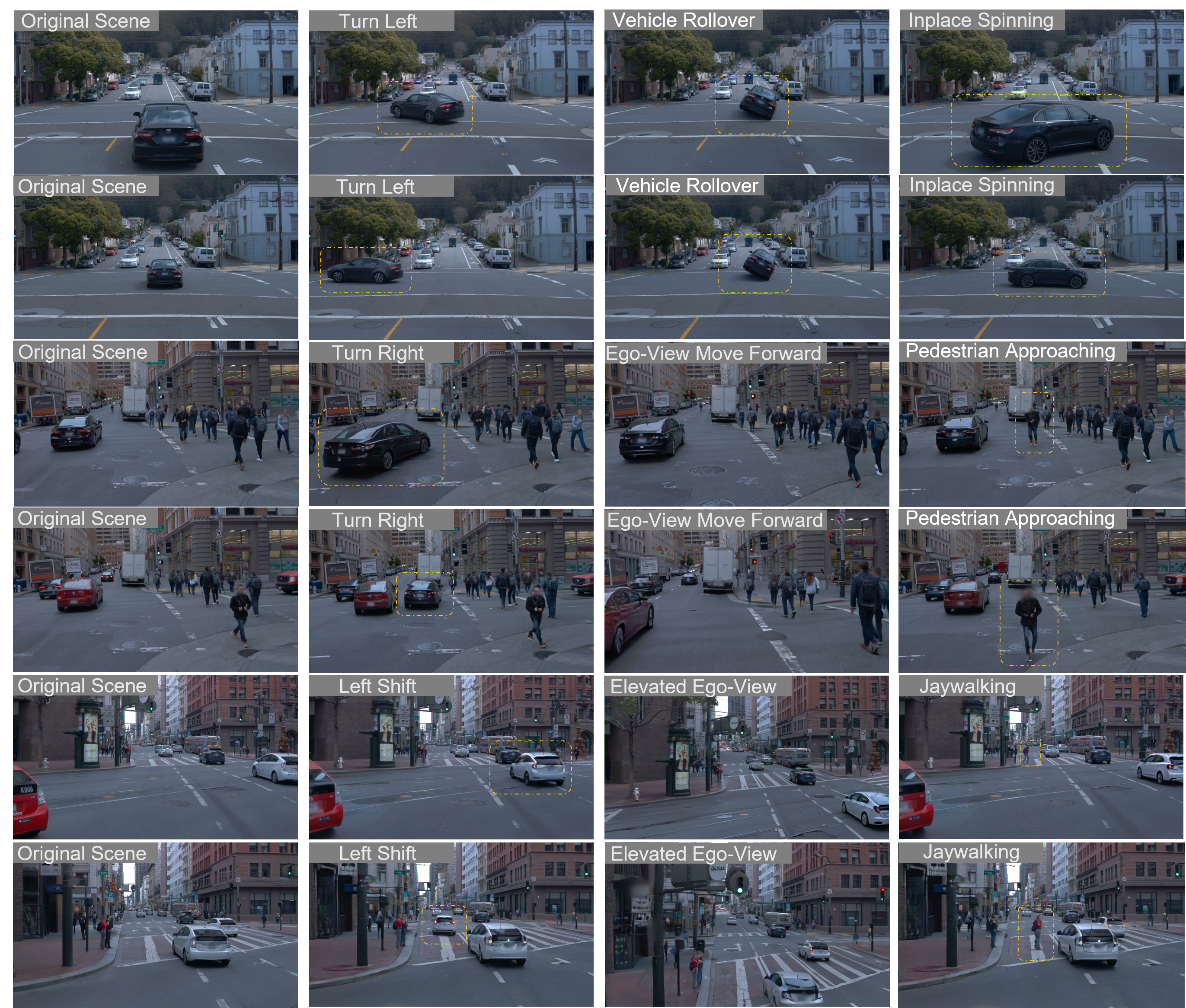}
    \vspace{-0.2in}
    \caption{\textbf{Qualitative demonstrations of ReinDriveGen on OOD safety-critical corner cases.} Every two consecutive rows show two frames from the same scene, with the edit type and edited region annotated in each image.}
    \label{fig:demo}
    \vspace{-0.2in}
\end{figure*}

\subsection{RL-based Post-training}
\label{sec:rl}
The supervised fine-tuning (SFT) stage described in Sec.~\ref{sec:pc_sim} trains the video diffusion model on pairs of
rendered pseudo-images and ground-truth videos, both captured under the original recorded ego-vehicle viewpoints and actor trajectories. At inference time, however, we wish to edit vehicle trajectories (\eg, simulating lead-vehicle collisions, drifting, spinning or other counterfactual maneuvers) or render from novel ego-vehicle viewpoints. In both cases, the rendered pseudo-image conditions deviate significantly from those seen during SFT, exhibiting novel occlusion patterns, unfamiliar vehicle orientations, and untextured completed surfaces, which creates a substantial domain gap.  


Empirically, we observe that vehicles are the primary source of visual degradation under these out-of-distribution (OOD) conditions, exhibiting distorted geometry, unrealistic textures, and implausible lighting, while static background regions remain largely unaffected. To address this limitation, we propose an RL-based post-training framework that directly optimizes vehicle generation quality under OOD conditions without requiring paired ground-truth supervision.

\subsubsection{Extending DiffusionNFT to Video Generation.}

We adopt DiffusionNFT~\cite{zheng2025diffusionnft} as our RL post-training framework, which achieves significantly faster training than FlowGRPO~\cite{liu2025flow} by performing policy optimization directly on the forward diffusion process in a classifier-free-guidance-free (CFG-free) manner. This efficiency advantage is particularly important for video diffusion RL, where the high dimensionality of spatiotemporal outputs makes both sampling and training substantially more expensive than in the image setting.

We extend DiffusionNFT from image to video generation. Let $\pi_\theta$ denote the video diffusion policy parameterized by $\theta$, and let $\mathbf{c}$ denote a condition consisting of a reference frame and a rendered pseudo-image sequence. We maintain a reference policy $\pi^{\mathrm{old}}$ (updated via EMA) alongside the current policy~$\pi_\theta$. In each iteration, we sample a batch of OOD conditions $\{\mathbf{c}_i\}$.
For each condition, the reference policy generates a group of $N$ candidate videos. A reward model scores each candidate, and the scores are used to optimize the following contrastive objective adapted from DiffusionNFT:
\vspace{-0.05in}
\begin{equation}
  \mathcal{L}(\theta)
  = \mathbb{E}_{\mathbf{c},\;
      \pi^{\mathrm{old}}(\mathbf{x}_0|\mathbf{c}),\; t}
    \Big[\,
      r\,\big\lVert
        v_\theta^{+}(\mathbf{x}_t,\mathbf{c},t)-\mathbf{v}
      \big\rVert_2^{2}
      +
      (1-r)\,\big\lVert
        v_\theta^{-}(\mathbf{x}_t,\mathbf{c},t)-\mathbf{v}
      \big\rVert_2^{2}
    \,\Big],
  \label{eq:nft}
\end{equation}
where $\mathbf{x}_0$ is a clean generated video sampled from $\pi^{\mathrm{old}}$, $\mathbf{x}_t$ is its noised version at
flow-matching timestep~$t$, $\mathbf{v}$ is the flow-matching velocity target, $r\!\in\![0,1]$ is the reward associated with $\mathbf{x}_0$, and $v_\theta^{+}$, $v_\theta^{-}$ are the implicit positive and negative velocity predictions as defined in~\cite{zheng2025diffusionnft}. Intuitively, for high-reward samples ($r\!\to\!1$) the loss encourages $v_\theta$ to reproduce similar high-quality outputs, while for low-reward samples ($r\!\to\!0$) it steers $v_\theta$ away from low-quality outputs.

\subsubsection{Pairwise Preference Model.}
\label{sec:reward}
A critical component of RL post-training is a reward model that evaluates the quality of vehicles in generated driving videos.
No existing reward model serves this purpose; general-purpose image quality or aesthetics models fail to capture domain-specific artifacts such as distorted vehicle geometry, unrealistic textures, or implausible lighting patterns. We therefore train a dedicated pairwise preference model to evaluate generated vehicle quality.

\paragraph{Preference data construction.}\quad
To train the pairwise preference model, we need paired data of relatively better and worse vehicle crops. We exploit a key empirical observation: when generating videos with the SFT model, subsampling the pseudo-image conditioning (\ie, providing a rendered pseudo-image every $k$-th frame instead of every frame) progressively degrades the visual quality of vehicles in the generated output.  This provides a natural and controllable mechanism for constructing preference pairs without manual annotation. Concretely, for each scene in the Waymo training dataset, we generate three versions of the output video using the SFT model with pseudo-image conditioning provided at
\emph{(i)}~every frame,
\emph{(ii)}~every 2nd frame, and
\emph{(iii)}~every 4th frame.
Since we focus exclusively on vehicle quality, we apply YOLO26~\cite{sapkota2025yolo26} to detect and crop all vehicle instances from each generated frame.  For a given vehicle instance in a given frame, the crops from different conditioning frequencies form naturally ordered
preference pairs:
(i)$\succ$(ii),\;
(ii)$\succ$(iii),\; and
(i)$\succ$(iii),
all paired on the same vehicle, same frame, and same scene, ensuring that the vehicle quality is the only varying factor.

To further diversify the degradation patterns and improve robustness, we augment both positive and negative samples with motion blur and elastic distortion, where the degradation strength applied to negative samples is always strictly greater than that applied to positive samples. This asymmetric augmentation ensures that the reward model learns to discriminate between relatively better and worse samples even when both exhibit significant artifacts. This property is essential for the early stages of RL training, where the policy generates low-quality outputs and the reward model must still provide a meaningful ranking among candidates to guide optimization. Further details are provided in the supplementary material.

\vspace{-0.2in}
\begin{figure*}[t]
    \centering
    \includegraphics[width=\linewidth]{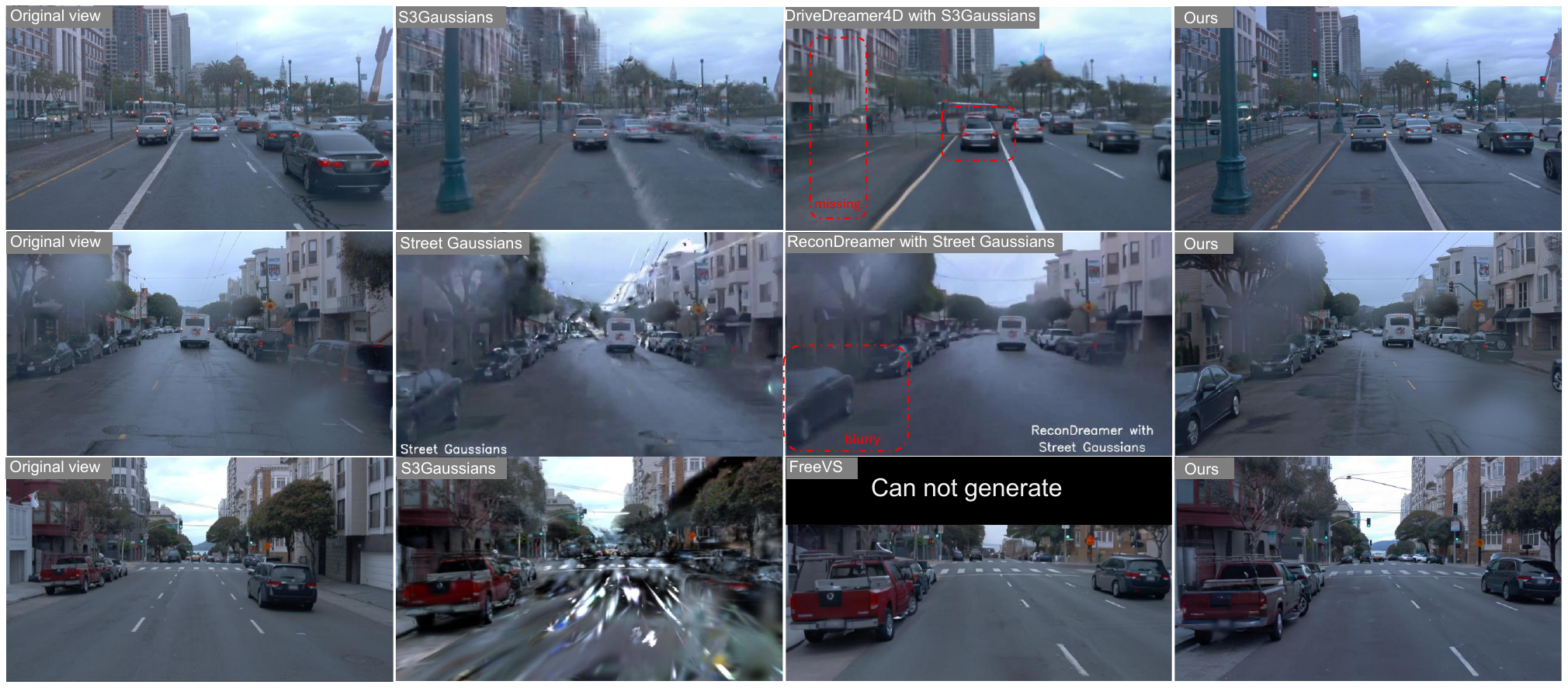}
    \vspace{-0.2in}
    \caption{Qualitative comparison of novel trajectories in the lane-change scenario.}
    \label{fig:comparison1}
    \vspace{-0.2in}
\end{figure*}

\paragraph{Model architecture.}\quad
Given two vehicle crops $I_1$ and $I_2$, we extract their CLS-token representations using a shared DINOv3~\cite{simeoni2025dinov3} ViT-H+ backbone~$f$, yielding $\mathbf{z}_1 = f(I_1)$ and $\mathbf{z}_2 = f(I_2)$ with
$\mathbf{z}_1, \mathbf{z}_2 \in \mathbb{R}^d$.  The two feature vectors are concatenated in both orders and passed through a shared MLP head~$h$ to produce scalar logits $s_{12} = h([\mathbf{z}_1;\,\mathbf{z}_2])$ and
$s_{21} = h([\mathbf{z}_2;\,\mathbf{z}_1])$.  The preference probability is then defined via an antisymmetric formulation:
\begin{equation}
  P(I_1 \succ I_2)
  = \sigma\!\Bigl(\frac{s_{12} - s_{21}}{2}\Bigr),
  \label{eq:pref}
\end{equation}
where $\sigma$ denotes the sigmoid function.  This design guarantees logical consistency, \ie,
$P(I_1\!\succ\!I_2) = 1 - P(I_2\!\succ\!I_1)$, since $\sigma(-x) = 1 - \sigma(x)$. The pairwise preference model is trained with binary cross-entropy loss on the constructed preference pairs.

\begin{figure*}[t]
    \centering
    \includegraphics[width=\linewidth]{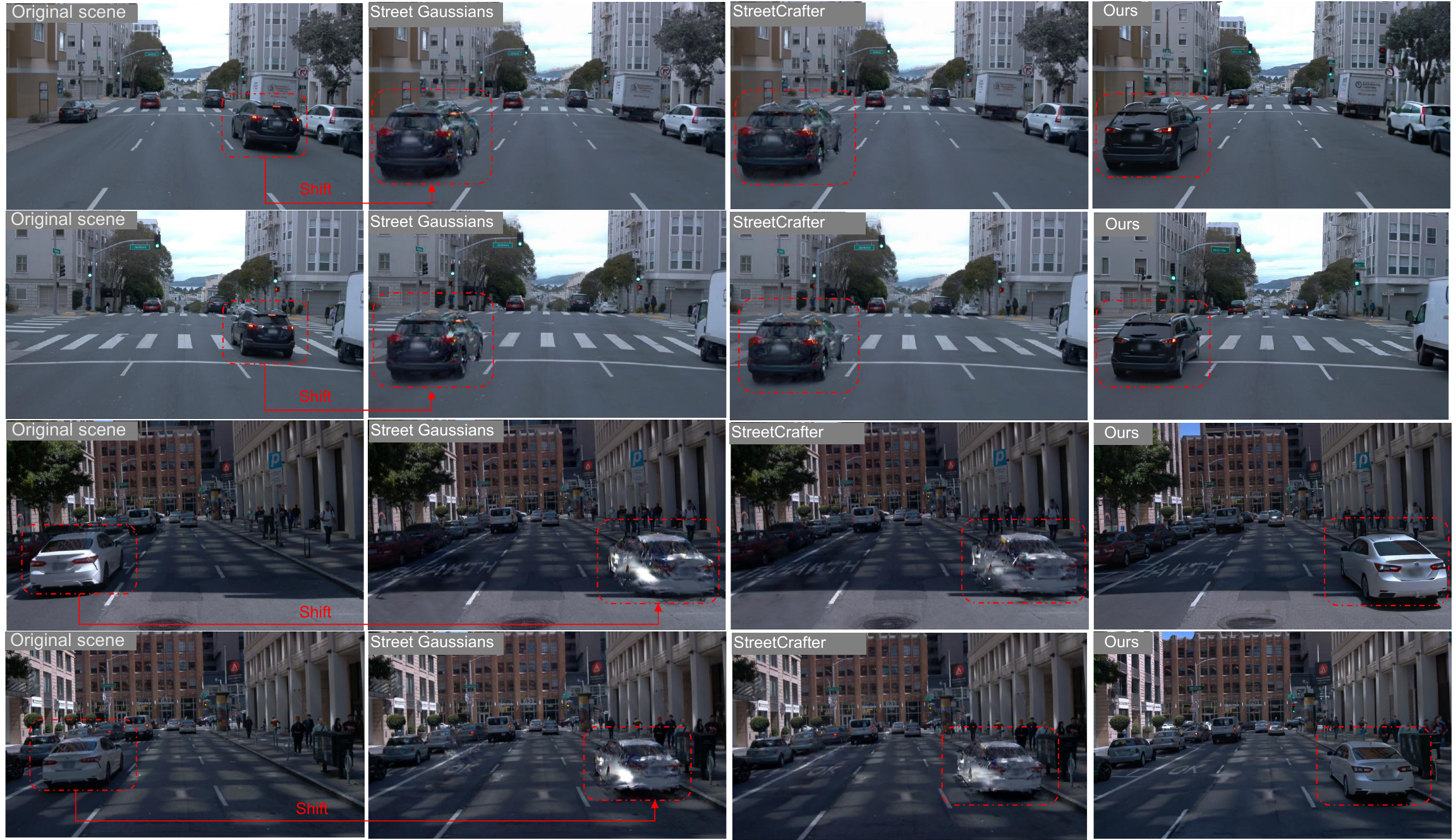}
    \vspace{-0.2in}
    \caption{Qualitative comparison of vehicle trajectory editing. Top two rows: the target vehicle is shifted 6m left and 4m backward. Bottom two rows: the target vehicle is shifted 6m to the right and 1m backward.}
    \label{fig:comparison2}
    \vspace{-0.2in}
\end{figure*}

\subsubsection{Pairwise Reward Mechanism.}\quad
Given the trained pairwise preference model, we propose a pairwise reward mechanism to aggregate pairwise comparison outcomes into per-candidate rewards within each group, providing robust training signals for RL-based post-training.

For each OOD condition~$\mathbf{c}$, we generate $N\!=\!16$ candidate videos $\{\mathbf{x}^1,\ldots,\mathbf{x}^N\}$.  Each candidate video is decomposed into individual frames.  We run YOLO26~\cite{sapkota2025yolo26} on the first candidate to detect and localize all vehicles in every frame, yielding a set of bounding boxes.  These bounding boxes are then applied identically to all $N$ candidates, so that exactly the same spatial region is cropped across candidates for fair comparison.  We treat each (frame index, bounding box) tuple as an independent evaluation unit; let $M$ denote the total number of such units across all frames.  For each unit~$m$, we crop the corresponding region from every candidate, producing $N$ crops $\{a_m^1,\ldots,a_m^N\}$.  We then perform all $\binom{N}{2}$ pairwise comparisons using our pairwise preference model (Eq.~\ref{eq:pref}).  For each pair $(a_m^i,a_m^j)$, the model outputs a preference probability $P(a_m^i\!\succ\!a_m^j)$.  The per-unit reward for candidate~$i$ at unit~$m$ is defined as its win rate:
\begin{equation}
  R_m^i
  = \frac{1}{N-1}\sum_{j\neq i}
    \mathds{1}\!\Big[P(a_m^i\!\succ\!a_m^j)>\tau\Big],
  \label{eq:winrate}
\end{equation}
where $\tau\!=\!0.85$ is a confidence threshold that filters out ambiguous comparisons.  The video-level reward $\hat{R}(\mathbf{x}^i)$ is computed as an area-weighted average of the per-unit win rates across all $M$ units, where each unit is weighted by its bounding-box pixel area so that artifacts on vehicles with larger area are prioritized.  Compared to pointwise scores, pairwise win rates span a wider and more discriminative range and are inherently robust to systematic biases in the reward model.  The resulting $\hat{R}(\mathbf{x}^i)$ is used in place of the pointwise reward~$r$ in the DiffusionNFT loss (Eq.~\ref{eq:nft}), which is seamless as DiffusionNFT only requires a scalar reward per sample. 

With the pairwise reward mechanism in place, we can now perform RL post-training on OOD conditions, such as lead-vehicle collisions, drifting, spinning, and counterfactual trajectories that place vehicles in extreme orientations. In each iteration, the reference policy generates $N\!=\!16$ candidate videos per condition, the pairwise reward mechanism produces a scalar reward~$\hat{R}(\mathbf{x}^i)$ for each candidate, and the policy is updated via the DiffusionNFT objective (Eq.~\ref{eq:nft}) to reinforce high-win-rate generation patterns while suppressing low-quality ones.  This enables the model to progressively improve vehicle quality under OOD conditions without any paired ground-truth supervision.

\section{Experiment}

\subsection{Demonstration of Actor and Viewpoint Manipulation}
Fig.~\ref{fig:demo} demonstrates that our method can manipulate vehicles, pedestrians, and cyclists, including maneuvers outside the training distribution such as in-place spinning, turning, vehicle rollover, and laterally shifted ego-vehicle viewpoints. We encourage the reader to refer to the supplementary material for video results.



\begin{table*}
\centering
\caption{Quantitative comparison in a lane-change scenario where the trajectory gradually shifts 4 m to the left.}
\vspace{-0.1in}
\resizebox{\textwidth}{!}{
\begin{tabular}{l|ccc|ccc|ccc|c}
\toprule
Models &  PVG & \makecell{DD4D w/.\\ PVG} & \makecell{Recon. w/. \\  PVG} & $S^3$Gauss. & \makecell{DD4D w/.\\  $S^3$Gauss.} & \makecell{Recon. w/. \\  $S^3$Gauss.} & Deform.-GS & \makecell{DD4D w/.\\ Deform.-GS} & \makecell{Recon. w/. \\  Deform.-GS} & Ours \\
\midrule
NTA-IoU$\uparrow$ & 0.256 & 0.438  & 0.464 & 0.175 & 0.495 & 0.413 & 0.240 & 0.335 & 0.443 & {\bf 0.549} \\
NTL-IoU$\uparrow$  & 50.70 & 53.06 & 53.21 & 49.05 & 53.42 & 51.62 & 51.62 & 52.93 & 53.78 & {\bf 56.13}  \\
FID$\downarrow$  & 105.29 & 71.52 & 74.32 & 124.90 & 66.93 & 123.61 & 92.24 & 77.32 & 76.24 & {\bf 51.99}\\

\bottomrule
\end{tabular}}

\label{tab:table_comparison1}
\end{table*}
\vspace{-0.3in}

\subsection{Implementation Details}

\subsubsection{Conditional Video Diffusion Model.}
We fine-tune the video diffusion model on 1{,}000 scenes from the Waymo dataset~\cite{Sun_2020_CVPR}. Each scene comprises five camera views; we use only the front, front-left, and front-right views for training. Starting from the VACE-1.3B model~\cite{jiang2025vace}, we fine-tune for approximately 500 H100 GPU hours with a batch size of 32, using the Adam optimizer with a constant learning rate of $1\!\times\!10^{-5}$.

\subsubsection{RL-Based Post-Training.}
We curate 20 representative OOD conditions covering in-place spinning, left/right turning, vehicle rollover, and laterally shifted ego-vehicle viewpoints. For each condition, we generate a group of $N=16$ candidate videos per rollout for policy optimization. To ensure training stability, we adopt LoRA for RL-based fine-tuning rather than updating full model weights. We use the Adam optimizer with a constant learning rate of $1\!\times\!10^{-5}$. All other hyperparameters of the DiffusionNFT framework, including $\beta$ and the data collection policy update strategy, follow the default settings in~\cite{zheng2025diffusionnft}. Post-training takes approximately 400 H100 GPU hours over 500 optimization steps.

For implementation details of the vehicle completion model and the pairwise preference model, please refer to the supplementary material.

\begin{figure}
    \centering
    \includegraphics[width=\linewidth]{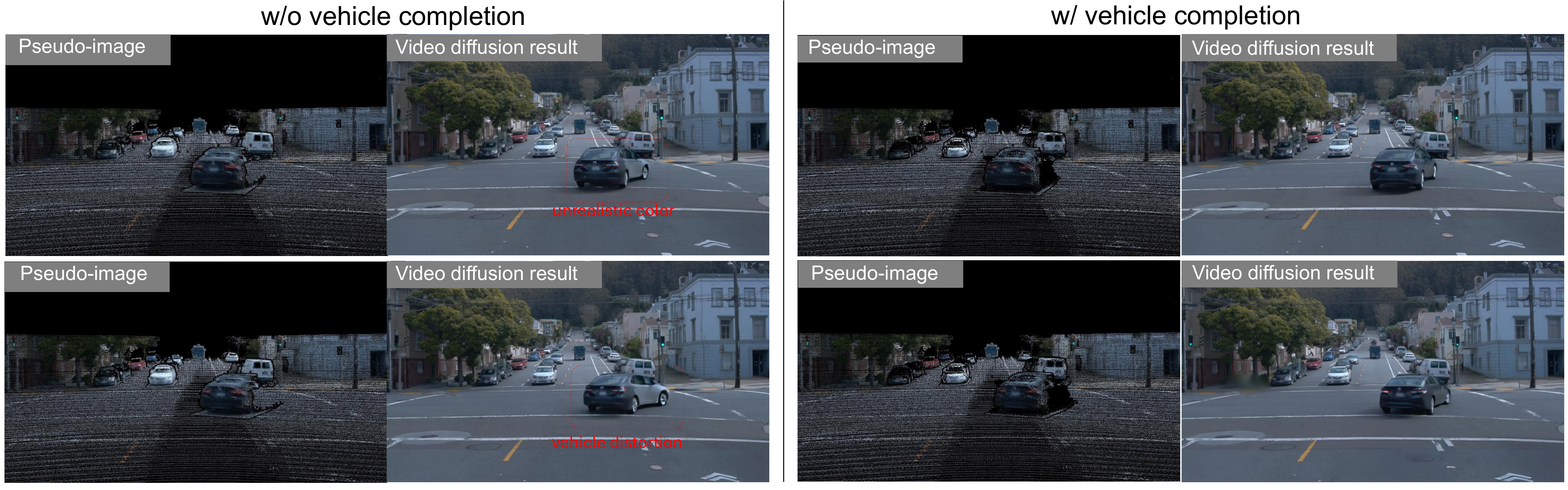}
    \vspace{-0.2in}
    \caption{\textbf{Ablation study on vehicle completion.} Vehicle completion fills in unobserved surfaces in the pseudo-images, effectively eliminating vehicle distortion and texture artifacts in the diffusion outputs.}
    \label{fig:ablation_completion}
\end{figure}

\vspace{-0.2in}

\subsection{Comparison}
We compare our method against existing approaches under two evaluation settings: novel ego-vehicle trajectory and vehicle trajectory editing.

\subsubsection{Novel Ego-Vehicle Trajectory.}
We evaluate on the off-trajectory benchmark proposed by DriveDreamer4D~\cite{zhao2024drive}, which introduces Novel Trajectory Agent IoU (NTA-IoU) and Novel Trajectory Lane IoU (NTL-IoU) to measure spatial consistency of vehicles and lane boundaries under novel viewpoints. To simulate a realistic lane-change scenario, we apply a lateral shift of 0.1\,m per frame over 40 frames, yielding a cumulative lateral shift of 4\,m. We compare with PVG~\cite{chen2026periodic}, DriveDreamer4D~\cite{zhao2024drive}, ReconDreamer~\cite{ni2025recondreamer}, S3Gaussian~\cite{huang2024s3gaussian}, and Deformable-GS~\cite{yang2023deformable3dgs}. We additionally report FID between original-trajectory and novel-trajectory images to evaluate photorealism. As shown in Tab.~\ref{tab:table_comparison1}, our method achieves the best performance across all metrics. The qualitative comparison in Fig.~\ref{fig:comparison1} is consistent with the quantitative results: our method produces the most photorealistic outputs. Since FreeVS~\cite{wang2024freevs} only generates partial novel views, we include it in qualitative comparisons only.


    

\begin{table}[t]
  \centering
  \caption{Quantitative comparison on the vehicle trajectory editing scenario with a 5m lateral shift.}
  \vspace{-0.1in}
  \label{tab:quantitative_comparison2}
  \setlength{\textwidth}{3pt} 

  \begin{tabular}{l|c|c|c|c}
    \hline  
    Method & Motion smoothness$\uparrow$ & Background consistency$\uparrow$ & Image quality$\uparrow$ & FID$\downarrow$ \\
    \hline 
    Street Gauss. & 0.981111 & 0.930975 & 0.639775 & 84.7970 \\
    StreetCrafter & {\bf 0.981715} & 0.906872 & 0.650856 & 83.6278 \\
    Ours          & 0.977355 & {\bf 0.939697} & {\bf 0.692809} & {\bf 80.6267} \\
    \hline 
  \end{tabular}
  \vspace{-0.2in}
\end{table}

\begin{figure}
    \centering
    \includegraphics[width=\linewidth]{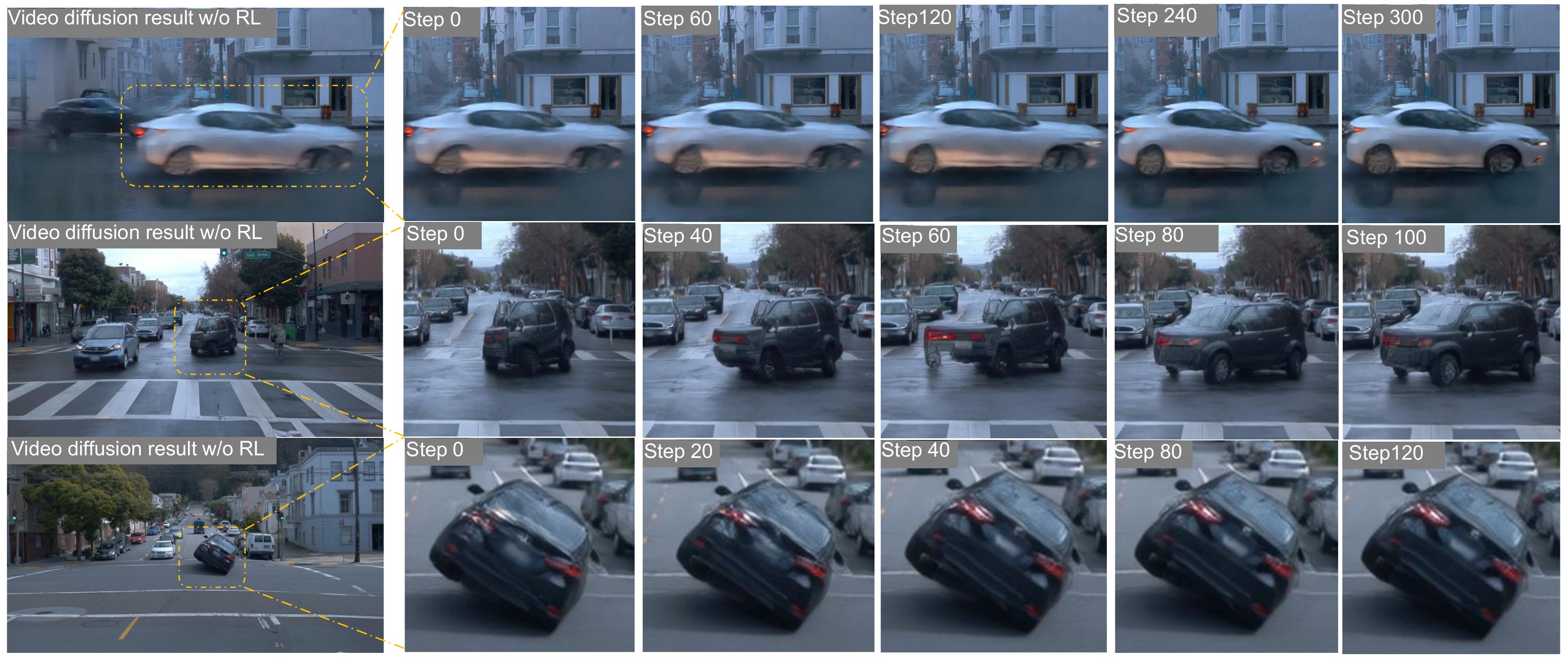}
    \vspace{-0.2in}
    \caption{\textbf{Ablation study on RL-based post-training.} From top to bottom: doubled vehicle speed, in-place spinning, and vehicle rollover. As the RL training step increases, the generated vehicles exhibit progressively better geometry, texture fidelity, and lighting plausibility.}
    \label{fig:ablation}
    \vspace{-0.2in}
\end{figure}

\subsubsection{Vehicle Trajectory Editing.}
Since several recent methods, including DriveDreamer4D~\cite{zhao2024drive}, ReconDreamer~\cite{ni2025recondreamer}, and FreeSim~\cite{fan2025freesim}, have not released their source code, we compare with Street Gaussians~\cite{yan2024street} and StreetCrafter~\cite{yan2025streetcrafter} on this setting. We keep the ego-vehicle trajectory fixed and manually edit a target vehicle's trajectory to evaluate each method's ability to handle modified scenes. The qualitative comparison is shown in Fig.~\ref{fig:comparison2}. Street Gaussians and StreetCrafter fail to synthesize plausible content in regions that are unobserved in the original recorded views and produce incorrect lighting on the edited vehicles. In contrast, our method successfully generates realistic content for invisible regions from the recorded views and produces photorealistic lighting. For quantitative evaluation, we adopt three metrics from VBench~\cite{huang2023vbench}, namely image quality, background consistency, and motion smoothness, and evaluate on the 5 scenes released by StreetCrafter with a 5\,m lateral vehicle shift. As shown in Tab.~\ref{tab:quantitative_comparison2}, our method achieves the best performance in FID, image quality, and background consistency.

\begin{figure*}
    \centering
    \includegraphics[width=\linewidth]{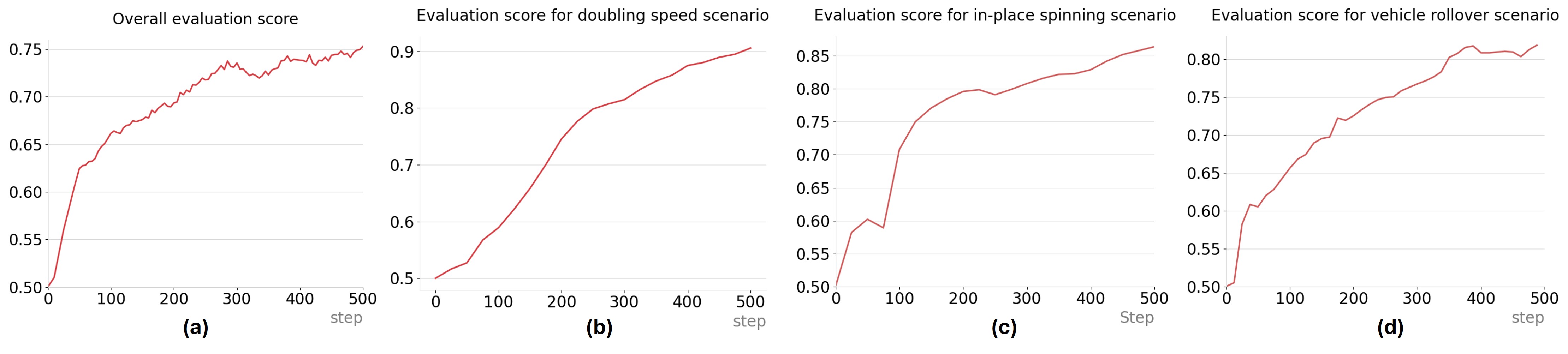}
    \vspace{-0.3in}
    \caption{\textbf{Evaluation score curves during RL training.} (a) Overall evaluation score averaged across all OOD conditions. (b)--(d) Per-scenario scores for doubling speed, in-place spinning, and vehicle rollover, respectively.}
    \label{fig:curve_reward}
    \vspace{-0.3in}
\end{figure*}

\subsection{Ablation Studies}
We conduct ablation studies to validate the effectiveness of two key components of our framework: vehicle point cloud completion (Sec.~\ref{sec:pc_sim}) and RL-based post-training (Sec.~\ref{sec:rl}).

\subsubsection{Effect of Vehicle Completion.}
To isolate the contribution of vehicle completion, we evaluate on a vehicle trajectory editing scenario where a target vehicle is repositioned along a modified path, exposing surfaces that are unobserved in the original recording. This setting is specifically chosen because repositioned vehicles are viewed from novel angles, making geometric completeness critical. Fig.~\ref{fig:ablation_completion} presents the rendered pseudo-images with and without vehicle completion, along with the corresponding video diffusion outputs. With vehicle completion, the pseudo-images provide geometrically complete vehicle silhouettes, enabling the diffusion model to produce vehicles with consistent geometry and plausible textures. Without it, large holes on occluded surfaces lead to noticeable vehicle distortion and severe texture artifacts, confirming that vehicle completion is essential for high-quality generation under trajectory edits.

\subsubsection{Effect of RL-Based Post-Training.}
\label{sec:ablation_rl}
Since the pairwise reward mechanism (Eq.~\ref{eq:winrate}) computes win rates within each group of $N$ candidates sampled for a given condition, the resulting rewards reflect only relative quality rankings within that group. They are not comparable across different groups. We therefore design a dedicated evaluation method that is executed every 5 training steps. Specifically, we fix the random seed to 0 to eliminate stochasticity, and for each OOD condition~$\mathbf{c}$, we generate one video from the SFT baseline $\pi_{\mathrm{sft}}$ and one from the current RL-trained model $\pi_{\theta_k}$ at step~$k$, both under identical conditions and the same seed. We then apply the same vehicle detection and cropping procedure described in Sec.~\ref{sec:reward} to obtain $M$ evaluation units. For each unit~$m$, we compute the pairwise preference probability $P(a_m^{\mathrm{rl}}\!\succ\!a_m^{\mathrm{sft}})$ using our pairwise preference model . The evaluation score for a single condition is defined as the area-weighted average of these preference probabilities across all $M$ units:
\begin{equation}
  E(\mathbf{c})
  = \frac{\sum_{m=1}^{M} w_m \cdot P(a_m^{\mathrm{rl}}\!\succ\!a_m^{\mathrm{sft}})}
         {\sum_{m=1}^{M} w_m},
  \label{eq:eval}
\end{equation}
where $w_m$ is the bounding-box pixel area of unit~$m$. The overall evaluation score is then obtained by averaging $E(\mathbf{c})$ across all evaluation conditions. At step~0, the RL model is identical to the SFT baseline, so both models produce the same video, yielding $P(a_m^{\mathrm{rl}}\!\succ\!a_m^{\mathrm{sft}})=0.5$ for all units and hence an overall score of $0.5$. A score rising above $0.5$ indicates that the RL-trained model generates higher-quality vehicles than the SFT baseline.

Fig.~\ref{fig:curve_reward}\,(a) shows the overall evaluation score over the course of RL training, demonstrating a steady improvement as training progresses. We additionally present three representative OOD scenarios in Fig.~\ref{fig:ablation}, with all visualizations generated using a fixed seed of 0 for fair comparison across different training steps. From top to bottom, the three scenarios are: (1)~doubling the original vehicle speed, (2)~in-place vehicle spinning, and (3)~vehicle rollover. The visual results clearly show a progressive refinement in vehicle geometry, texture fidelity, and lighting plausibility as RL training proceeds, confirming that the post-training framework effectively improves generation quality under OOD conditions. The per-scenario evaluation scores are reported in Fig.~\ref{fig:curve_reward}\,(b),\,(c), and\,(d), respectively, all exhibiting consistent upward trends that corroborate the qualitative observations.

\subsection{Limitations and Conclusion}
\label{sec:conclusion}

We have presented ReinDriveGen, a unified framework for photorealistic and freely editable driving scene simulation that combines explicit 3D point cloud editing with diffusion-based video synthesis. A vehicle completion module enables free-form manipulation of vehicle trajectories without geometric artifacts, and an RL-based post-training strategy with a pairwise reward mechanism substantially improves vehicle quality under out-of-distribution scenarios. Currently, our method is constrained by GPU memory to generating 49-frame clips, with each clip taking approximately one minute, which does not yet support real-time applications.

More broadly, our work demonstrates that RL-based post-training can effectively improve diffusion model performance beyond the coverage of supervised training data. In autonomous driving, safety-critical corner cases, such as in-place spinning and vehicle rollover, are inherently rare, making the resulting distribution gap a fundamental challenge rather than a data collection problem. The proposed approach offers a principled solution to this challenge, and we believe the same paradigm can be extended to other domains where training data cannot fully cover the space of desired test-time conditions.

%
\bibliographystyle{splncs04}
\bibliography{main}

@String(CVPR  = {IEEE Conf. Comput. Vis. Pattern Recog.})

@String(ECCV  = {Eur. Conf. Comput. Vis.})

@String(NeurIPS = {Adv. Neural Inform. Process. Syst.})

@String(CVPR  = {CVPR})

@String(ECCV  = {ECCV})

@String(NeurIPS = {NeurIPS})

@inproceedings{gao2025magicdrive-v2,
  title={{MagicDrive-V2}: High-Resolution Long Video Generation for Autonomous Driving with Adaptive Control},
  author={Gao, Ruiyuan and Chen, Kai and Xiao, Bo and Hong, Lanqing and Li, Zhenguo and Xu, Qiang},
  booktitle={Proceedings of the IEEE/CVF International Conference on Computer Vision},
  year={2025}
}

@inproceedings{zhao2024drive,
    title={DriveDreamer4D: World Models Are Effective Data Machines for 4D Driving Scene Representation}, 
    author={Guosheng Zhao and Chaojun Ni and Xiaofeng Wang and Zheng Zhu and Xueyang Zhang and Yida Wang and Guan Huang and Xinze Chen and Boyuan Wang and Youyi Zhang and Wenjun Mei and Xingang Wang},
    journal={arxiv arXiv preprint arXiv:2410.13571},
    year={2024},
}

@article{ni2025maskgwm,
  title={MaskGWM: A Generalizable Driving World Model with Video Mask Reconstruction},
  author={Ni, Jingcheng and Guo, Yuxin and Liu, Yichen and Chen, Rui and Lu, Lewei and Wu, Zehuan},
  journal={arXiv preprint arXiv:2502.11663},
  year={2025}
}

@misc{nvidia2026worldsimulationvideofoundation,
      title={World Simulation with Video Foundation Models for Physical AI}, 
      author={NVIDIA and : and Arslan Ali and Junjie Bai and Maciej Bala and Yogesh Balaji and Aaron Blakeman and Tiffany Cai and Jiaxin Cao and Tianshi Cao and Elizabeth Cha and Yu-Wei Chao and Prithvijit Chattopadhyay and Mike Chen and Yongxin Chen and Yu Chen and Shuai Cheng and Yin Cui and Jenna Diamond and Yifan Ding and Jiaojiao Fan and Linxi Fan and Liang Feng and Francesco Ferroni and Sanja Fidler and Xiao Fu and Ruiyuan Gao and Yunhao Ge and Jinwei Gu and Aryaman Gupta and Siddharth Gururani and Imad El Hanafi and Ali Hassani and Zekun Hao and Jacob Huffman and Joel Jang and Pooya Jannaty and Jan Kautz and Grace Lam and Xuan Li and Zhaoshuo Li and Maosheng Liao and Chen-Hsuan Lin and Tsung-Yi Lin and Yen-Chen Lin and Huan Ling and Ming-Yu Liu and Xian Liu and Yifan Lu and Alice Luo and Qianli Ma and Hanzi Mao and Kaichun Mo and Seungjun Nah and Yashraj Narang and Abhijeet Panaskar and Lindsey Pavao and Trung Pham and Morteza Ramezanali and Fitsum Reda and Scott Reed and Xuanchi Ren and Haonan Shao and Yue Shen and Stella Shi and Shuran Song and Bartosz Stefaniak and Shangkun Sun and Shitao Tang and Sameena Tasmeen and Lyne Tchapmi and Wei-Cheng Tseng and Jibin Varghese and Andrew Z. Wang and Hao Wang and Haoxiang Wang and Heng Wang and Ting-Chun Wang and Fangyin Wei and Jiashu Xu and Dinghao Yang and Xiaodong Yang and Haotian Ye and Seonghyeon Ye and Xiaohui Zeng and Jing Zhang and Qinsheng Zhang and Kaiwen Zheng and Andrew Zhu and Yuke Zhu},
      year={2026},
      eprint={2511.00062},
      archivePrefix={arXiv},
      primaryClass={cs.CV},
      url={https://arxiv.org/abs/2511.00062}, 
}

@inproceedings{mildenhall2020nerf,
  title={NeRF: Representing Scenes as Neural Radiance Fields for View Synthesis},
  author={Ben Mildenhall and Pratul P. Srinivasan and Matthew Tancik and Jonathan T. Barron and Ravi Ramamoorthi and Ren Ng},
  year={2020},
  booktitle={ECCV},
}

@article{kerbl20233d,
  title={3d gaussian splatting for real-time radiance field rendering.},
  author={Kerbl, Bernhard and Kopanas, Georgios and Leimk{\"u}hler, Thomas and Drettakis, George and others},
  journal={ACM Trans. Graph.},
  volume={42},
  number={4},
  pages={139--1},
  year={2023}
}

@inproceedings{ni2025recondreamer,
  title={Recondreamer: Crafting world models for driving scene reconstruction via online restoration},
  author={Ni, Chaojun and Zhao, Guosheng and Wang, Xiaofeng and Zhu, Zheng and Qin, Wenkang and Huang, Guan and Liu, Chen and Chen, Yuyin and Wang, Yida and Zhang, Xueyang and others},
  booktitle={Proceedings of the Computer Vision and Pattern Recognition Conference},
  pages={1559--1569},
  year={2025}
}

@inproceedings{yan2025streetcrafter,
  title={Streetcrafter: Street view synthesis with controllable video diffusion models},
  author={Yan, Yunzhi and Xu, Zhen and Lin, Haotong and Jin, Haian and Guo, Haoyu and Wang, Yida and Zhan, Kun and Lang, Xianpeng and Bao, Hujun and Zhou, Xiaowei and others},
  booktitle={Proceedings of the Computer Vision and Pattern Recognition Conference},
  pages={822--832},
  year={2025}
}

@article{zheng2025diffusionnft,
  title={Diffusionnft: Online diffusion reinforcement with forward process},
  author={Zheng, Kaiwen and Chen, Huayu and Ye, Haotian and Wang, Haoxiang and Zhang, Qinsheng and Jiang, Kai and Su, Hang and Ermon, Stefano and Zhu, Jun and Liu, Ming-Yu},
  journal={arXiv preprint arXiv:2509.16117},
  year={2025}
}

@article{lipman2022flow,
  title={Flow matching for generative modeling},
  author={Lipman, Yaron and Chen, Ricky TQ and Ben-Hamu, Heli and Nickel, Maximilian and Le, Matt},
  journal={arXiv preprint arXiv:2210.02747},
  year={2022}
}

@article{Pref-GRPO&UniGenBench,
  title={Pref-GRPO: Pairwise Preference Reward-based GRPO for Stable Text-to-Image Reinforcement Learning},
  author={Wang, Yibin and Li, Zhimin and Zang, Yuhang and Zhou, Yujie and Bu, Jiazi and Wang, Chunyu and Lu, Qinglin and Jin, Cheng and Wang, Jiaqi},
  journal={arXiv preprint arXiv:2508.20751},
  year={2025}
}

@article{huang2024s3gaussian,
        title={S3Gaussian: Self-Supervised Street Gaussians for Autonomous Driving},
        author={Huang, Nan and Wei, Xiaobao and Zheng, Wenzhao and An, Pengju and Lu, Ming and Zhan, Wei and Tomizuka,    Masayoshi and Keutzer, Kurt and Zhang, Shanghang},
        journal={arXiv preprint arXiv:2405.20323},
        year={2024}
      }

@inproceedings{wu2023mars,
  title={Mars: An instance-aware, modular and realistic simulator for autonomous driving},
  author={Wu, Zirui and Liu, Tianyu and Luo, Liyi and Zhong, Zhide and Chen, Jianteng and Xiao, Hongmin and Hou, Chao and Lou, Haozhe and Chen, Yuantao and Yang, Runyi and others},
  booktitle={CAAI International Conference on Artificial Intelligence},
  pages={3--15},
  year={2023},
  organization={Springer}
}

@article{guo2023streetsurf,
  title={StreetSurf: Extending Multi-view Implicit Surface Reconstruction to Street Views},
  author={Guo, Jianfei and Deng, Nianchen and Li, Xinyang and Bai, Yeqi and Shi, Botian and Wang, Chiyu and Ding, Chenjing and Wang, Dongliang and Li, Yikang},
  journal={arXiv preprint arXiv:2306.04988},
  year={2023}
}

@inproceedings{turki2023suds,
  title={SUDS: Scalable Urban Dynamic Scenes},
  author={Turki, Haithem and Zhang, Jason Y and Ferroni, Francesco and Ramanan, Deva},
  booktitle={Proceedings of the IEEE/CVF Conference on Computer Vision and Pattern Recognition},
  pages={12375--12385},
  year={2023}
}

@inproceedings{yang2023unisim,
  title={UniSim: A Neural Closed-Loop Sensor Simulator},
  author={Yang, Ze and Chen, Yun and Wang, Jingkang and Manivasagam, Sivabalan and Ma, Wei-Chiu and Yang, Anqi Joyce and Urtasun, Raquel},
  booktitle={Proceedings of the IEEE/CVF Conference on Computer Vision and Pattern Recognition},
  pages={1389--1399},
  year={2023}
}

@inproceedings{tonderski2024neurad,
  title={Neurad: Neural rendering for autonomous driving},
  author={Tonderski, Adam and Lindstr{\"o}m, Carl and Hess, Georg and Ljungbergh, William and Svensson, Lennart and Petersson, Christoffer},
  booktitle={Proceedings of the IEEE/CVF Conference on Computer Vision and Pattern Recognition},
  pages={14895--14904},
  year={2024}
}

@inproceedings{zhou2024drivinggaussian,
  title={Drivinggaussian: Composite gaussian splatting for surrounding dynamic autonomous driving scenes},
  author={Zhou, Xiaoyu and Lin, Zhiwei and Shan, Xiaojun and Wang, Yongtao and Sun, Deqing and Yang, Ming-Hsuan},
  booktitle={Proceedings of the IEEE/CVF Conference on Computer Vision and Pattern Recognition},
  pages={21634--21643},
  year={2024}
}

@article{wang2024freevs,
  title={Freevs: Generative view synthesis on free driving trajectory},
  author={Wang, Qitai and Fan, Lue and Wang, Yuqi and Chen, Yuntao and Zhang, Zhaoxiang},
  journal={arXiv preprint arXiv:2410.18079},
  year={2024}
}

@InProceedings{Sun_2020_CVPR, author = {Sun, Pei and Kretzschmar, Henrik and Dotiwalla, Xerxes and Chouard, Aurelien and Patnaik, Vijaysai and Tsui, Paul and Guo, James and Zhou, Yin and Chai, Yuning and Caine, Benjamin and Vasudevan, Vijay and Han, Wei and Ngiam, Jiquan and Zhao, Hang and Timofeev, Aleksei and Ettinger, Scott and Krivokon, Maxim and Gao, Amy and Joshi, Aditya and Zhang, Yu and Shlens, Jonathon and Chen, Zhifeng and Anguelov, Dragomir}, title = {Scalability in Perception for Autonomous Driving: Waymo Open Dataset}, booktitle = {Proceedings of the IEEE/CVF Conference on Computer Vision and Pattern Recognition (CVPR)}, month = {June}, year = {2020} }

@inproceedings{jiang2025vace,
  title={Vace: All-in-one video creation and editing},
  author={Jiang, Zeyinzi and Han, Zhen and Mao, Chaojie and Zhang, Jingfeng and Pan, Yulin and Liu, Yu},
  booktitle={Proceedings of the IEEE/CVF International Conference on Computer Vision},
  pages={17191--17202},
  year={2025}
}

@inproceedings{yu2021pointr,
  title={Pointr: Diverse point cloud completion with geometry-aware transformers},
  author={Yu, Xumin and Rao, Yongming and Wang, Ziyi and Liu, Zuyan and Lu, Jiwen and Zhou, Jie},
  booktitle={Proceedings of the IEEE/CVF international conference on computer vision},
  pages={12498--12507},
  year={2021}
}

@article{zhang2026ga,
  title={GA-Drive: Geometry-Appearance Decoupled Modeling for Free-viewpoint Driving Scene Generatio},
  author={Zhang, Hao and Fan, Lue and Wang, Qitai and Li, Wenbo and Wu, Zehuan and Lu, Lewei and Zhang, Zhaoxiang and Li, Hongsheng},
  journal={arXiv preprint arXiv:2602.20673},
  year={2026}
}

@article{liu2025flow,
  title={Flow-grpo: Training flow matching models via online rl},
  author={Liu, Jie and Liu, Gongye and Liang, Jiajun and Li, Yangguang and Liu, Jiaheng and Wang, Xintao and Wan, Pengfei and Zhang, Di and Ouyang, Wanli},
  journal={arXiv preprint arXiv:2505.05470},
  year={2025}
}

@article{sapkota2025yolo26,
  title={YOLO26: key architectural enhancements and performance benchmarking for real-time object detection},
  author={Sapkota, Ranjan and Cheppally, Rahul Harsha and Sharda, Ajay and Karkee, Manoj},
  journal={arXiv preprint arXiv:2509.25164},
  year={2025}
}

@article{chen2026periodic,
  title={Periodic vibration gaussian: Dynamic urban scene reconstruction and real-time rendering},
  author={Chen, Yurui and Gu, Chun and Jiang, Junzhe and Zhu, Xiatian and Zhang, Li},
  journal={International Journal of Computer Vision},
  year={2026},
}

@article{yang2023deformable3dgs,
    title={Deformable 3D Gaussians for High-Fidelity Monocular Dynamic Scene Reconstruction},
    author={Yang, Ziyi and Gao, Xinyu and Zhou, Wen and Jiao, Shaohui and Zhang, Yuqing and Jin, Xiaogang},
    journal={arXiv preprint arXiv:2309.13101},
    year={2023}
}

@InProceedings{huang2023vbench,
     title={{VBench}: Comprehensive Benchmark Suite for Video Generative Models},
     author={Huang, Ziqi and He, Yinan and Yu, Jiashuo and Zhang, Fan and Si, Chenyang and Jiang, Yuming and Zhang, Yuanhan and Wu, Tianxing and Jin, Qingyang and Chanpaisit, Nattapol and Wang, Yaohui and Chen, Xinyuan and Wang, Limin and Lin, Dahua and Qiao, Yu and Liu, Ziwei},
     booktitle={Proceedings of the IEEE/CVF Conference on Computer Vision and Pattern Recognition},
     year={2024}
 }

@misc{simeoni2025dinov3,
  title={{DINOv3}},
  author={Sim{\'e}oni, Oriane and Vo, Huy V. and Seitzer, Maximilian and Baldassarre, Federico and Oquab, Maxime and Jose, Cijo and Khalidov, Vasil and Szafraniec, Marc and Yi, Seungeun and Ramamonjisoa, Micha{\"e}l and Massa, Francisco and Haziza, Daniel and Wehrstedt, Luca and Wang, Jianyuan and Darcet, Timoth{\'e}e and Moutakanni, Th{\'e}o and Sentana, Leonel and Roberts, Claire and Vedaldi, Andrea and Tolan, Jamie and Brandt, John and Couprie, Camille and Mairal, Julien and J{\'e}gou, Herv{\'e} and Labatut, Patrick and Bojanowski, Piotr},
  year={2025},
  eprint={2508.10104},
  archivePrefix={arXiv},
  primaryClass={cs.CV},
  url={https://arxiv.org/abs/2508.10104},
}

@inproceedings{gao2024vista,
 title={Vista: A Generalizable Driving World Model with High Fidelity and Versatile Controllability}, 
 author={Shenyuan Gao and Jiazhi Yang and Li Chen and Kashyap Chitta and Yihang Qiu and Andreas Geiger and Jun Zhang and Hongyang Li},
 booktitle={Advances in Neural Information Processing Systems (NeurIPS)},
 year={2024}
}

@inproceedings{fan2025freesim,
  title={Freesim: Toward free-viewpoint camera simulation in driving scenes},
  author={Fan, Lue and Zhang, Hao and Wang, Qitai and Li, Hongsheng and Zhang, Zhaoxiang},
  booktitle={Proceedings of the Computer Vision and Pattern Recognition Conference},
  pages={12004--12014},
  year={2025}
}

@inproceedings{yan2024street,
    title={Street Gaussians: Modeling Dynamic Urban Scenes with Gaussian Splatting}, 
    author={Yunzhi Yan and Haotong Lin and Chenxu Zhou and Weijie Wang and Haiyang Sun and Kun Zhan and Xianpeng Lang and Xiaowei Zhou and Sida Peng},
    booktitle={ECCV},
    year={2024}
}

@article{chang2015shapenet,
  title={Shapenet: An information-rich 3d model repository},
  author={Chang, Angel X and Funkhouser, Thomas and Guibas, Leonidas and Hanrahan, Pat and Huang, Qixing and Li, Zimo and Savarese, Silvio and Savva, Manolis and Song, Shuran and Su, Hao and others},
  journal={arXiv preprint arXiv:1512.03012},
  year={2015}
}

\newpage
\appendix

\section{Details of Point Completion Model}
We fine-tune a pre-trained AdaPoinTr~\cite{yu2021pointr} model on the car category of Projected ShapeNet-55~\cite{chang2015shapenet} for 600 epochs. To bridge the sim-to-real gap with real-world KITTI LiDAR data, we simulate single-sided sensor visibility using 16 viewpoints per CAD model and inject synthetic noise. The model takes 2,048 sampled points as input and predicts 8,192 complete points. During training, we apply multi-view random sampling and X/Z-axis mirroring for data augmentation. We optimize the standard AdaPoinTr architecture (6-layer encoder, 8-layer decoder) using AdamW with a batch size of 32, an initial learning rate of \(1 \times 10^{-4}\), and a weight decay of \(5 \times 10^{-4}\).

\section{Details of Pairwise Preference Model} 
To prevent the preference model from trivially relying on low-level image sharpness, we introduce an asymmetric adversarial augmentation strategy. Specifically, we degrade high-quality images using structural-preserving perturbations (e.g., motion blur, ghosting, Gaussian blur, and elastic transformations), while artificially sharpening low-quality images via unsharp masking alongside structural distortions. With the DINOv3 backbone frozen, we train the preference head for 15 epochs using the AdamW optimizer. We set the batch size to 32, the initial learning rate to \(1 \times 10^{-5}\), and the weight decay to \(0.05\). The learning rate is decayed following a cosine schedule with a \(10\%\) warmup ratio. The network is optimized using a Binary Cross-Entropy (BCE) loss applied to the symmetric difference of the pairwise logits.

\begin{figure}
    \centering
    \includegraphics[width=\linewidth]{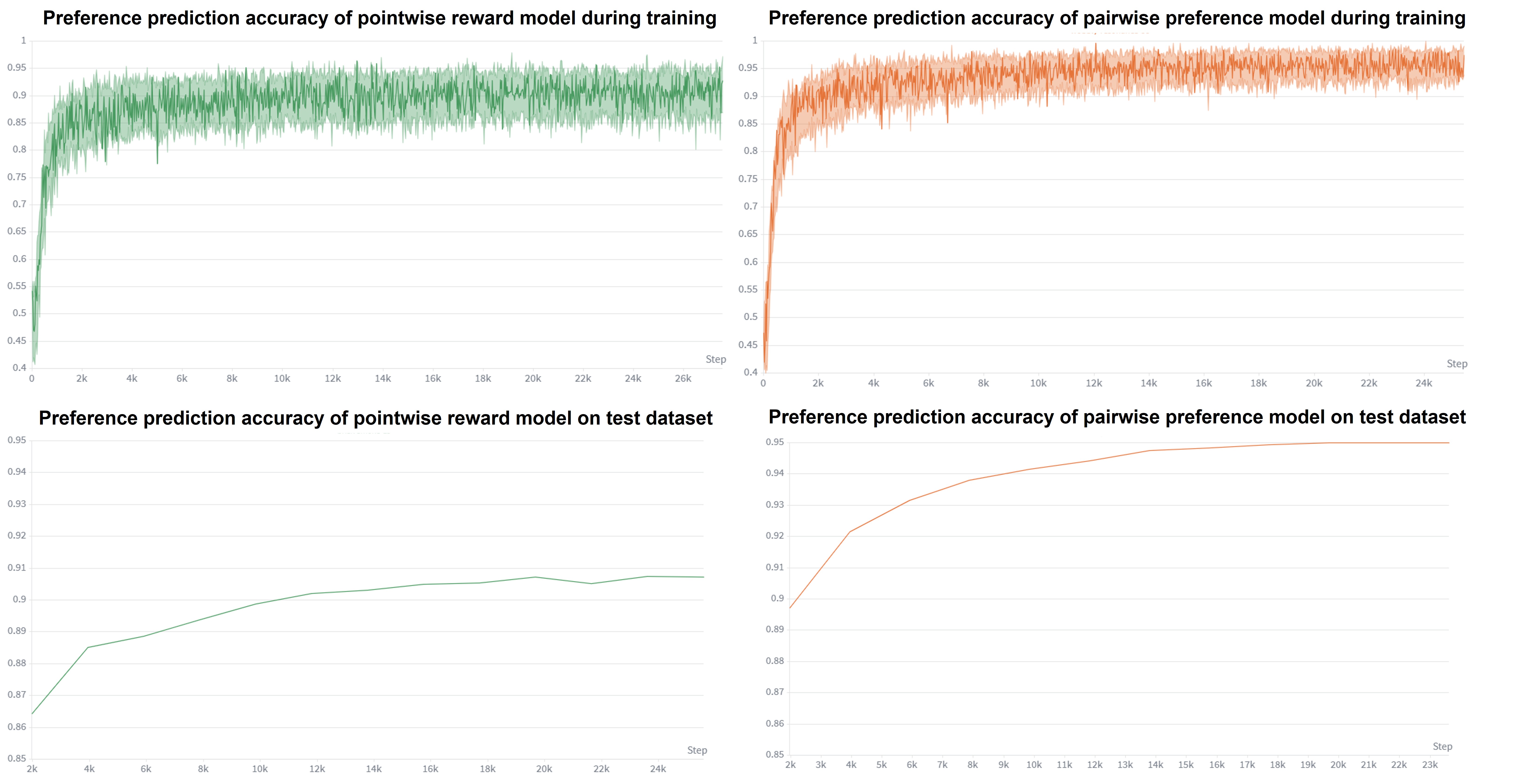}
    \vspace{-0.2in}
    \caption{Preference prediction accuracy of the pointwise reward model (green) and the pairwise preference model (orange) during training (top) and on the test set (bottom). The pairwise model achieves higher test-set accuracy (95\% vs.\ 91\%) with more stable training dynamics, demonstrating that jointly contrasting two samples within a shared forward pass yields more robust preference judgments than independent pointwise scoring.}
    \label{fig:pointwise_pairwise_curve}
    \vspace{-0.2in}
\end{figure}

\section{Ablation study on the Pairwise Reward model and the Pointwise Reward model}

\noindent\textbf{Pairwise vs.\ Pointwise Reward Model.}
A core design choice in our RL-based post-training is the use of a pairwise preference model (Sec.3.2) rather than a conventional pointwise reward model. To validate this choice, we train a pointwise baseline and compare both models in terms of preference prediction accuracy and downstream RL post-training quality.

\subsection{Pointwise reward model.}
The pointwise baseline shares the same frozen DINOv3~\cite{simeoni2025dinov3} ViT-H+ backbone as our pairwise model to extract CLS-token representations. Instead of jointly comparing two crops through the antisymmetric formulation, it attaches an independent MLP head $g$ that maps each CLS-token to a scalar quality score in $[0,1]$. The head architecture is: $\mathrm{Linear}(d, 1024) \!\to\! \mathrm{LN} \!\to\! \mathrm{GELU} \!\to\! \mathrm{Dropout}(0.1) \!\to\! \mathrm{Linear}(1024, 512) \!\to\! \mathrm{LN} \!\to\! \mathrm{GELU} \!\to\! \mathrm{Linear}(512, 1) \!\to\! \mathrm{Sigmoid}$, where $d$ is the backbone hidden dimension and $\mathrm{LN}$ denotes LayerNorm. Given two vehicle crops $I_1$ and $I_2$, the model independently computes scalar scores $r_1 = g(f(I_1))$ and $r_2 = g(f(I_2))$, and the preference probability under the Bradley-Terry framework is:
\begin{equation}
    P(I_1 \succ I_2) = \sigma(r_1 - r_2),
    \label{eq:bt}
\end{equation}
where $\sigma$ is the sigmoid function. Crucially, each image receives an absolute score in isolation, and the preference is determined solely by the difference of two independently computed scalars. The model is trained by maximizing the Bradley-Terry log-likelihood: $\mathcal{L}_{\mathrm{pt}} = -\log \sigma(r_{\mathrm{pos}} - r_{\mathrm{neg}})$.

\subsection{Preference accuracy comparison.}
Fig.~\ref{fig:pointwise_pairwise_curve} presents the preference prediction accuracy of both models on the training and test sets. The pairwise model consistently achieves higher test-set accuracy with more stable training dynamics, confirming that directly contrasting two samples within a shared forward pass yields more robust preference judgments than assigning absolute scores independently.

\begin{figure}
    \centering
    \includegraphics[width=\linewidth]{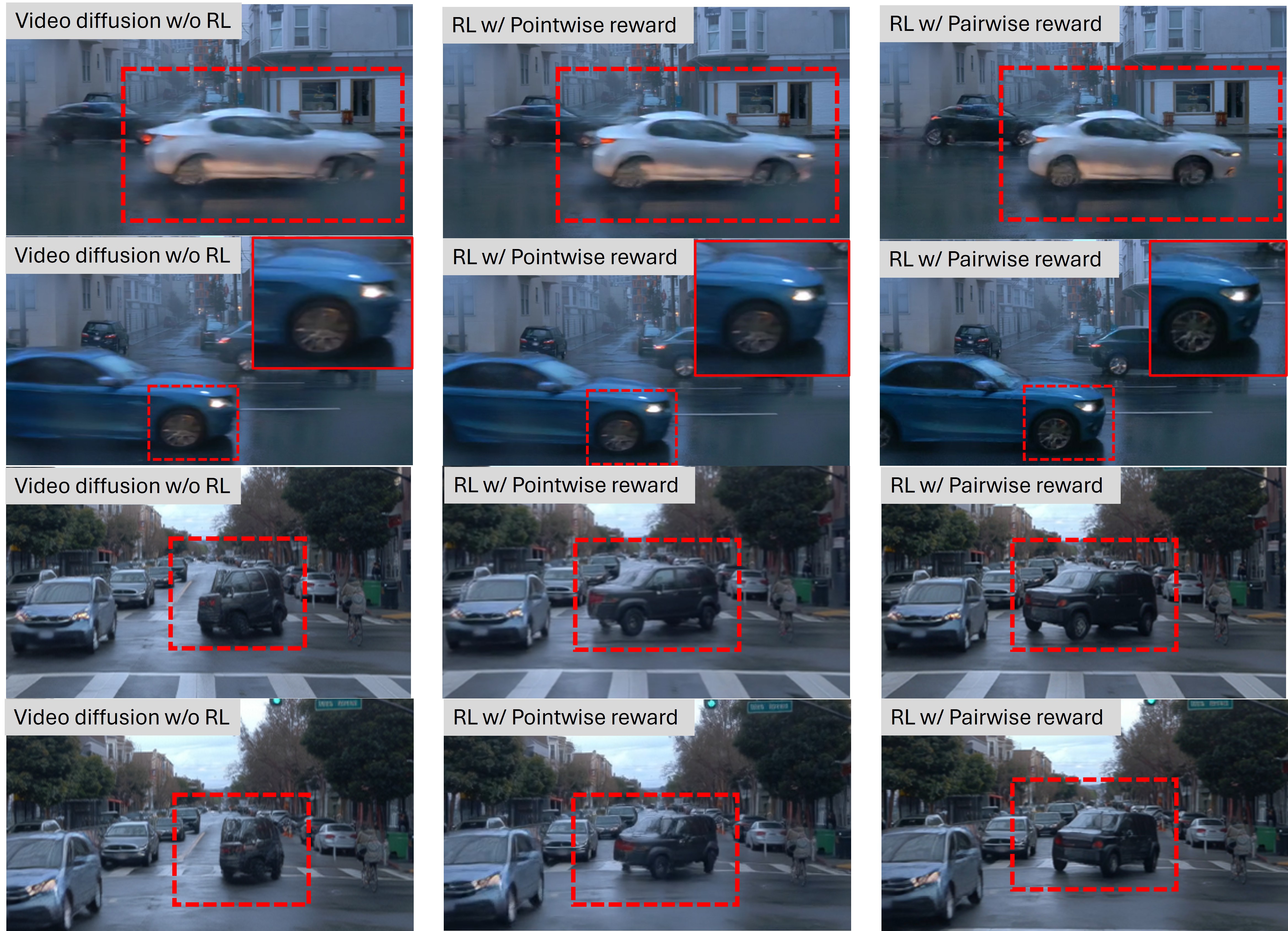}
    \vspace{-0.2in}
    \caption{\textbf{Qualitative comparison of RL post-training with pointwise vs.\ pairwise reward.} From left to right: video diffusion without RL, RL with pointwise reward, and RL with pairwise reward. Red dashed boxes highlight vehicle regions. While pointwise reward offers limited improvement and sometimes suffers from reward hacking artifacts, the pairwise reward mechanism consistently produces vehicles with better geometry, texture fidelity, and lighting plausibility, demonstrating its superiority as the reward signal for RL post-training.}
    \label{fig:pointwise_pairwise_comparison}
    \vspace{-0.2in}
\end{figure}

\subsection{RL post-training comparison.}
We substitute the pairwise reward in our RL pipeline with the pointwise model: each candidate's reward is its independent scalar score $r_i = g(f(I_i))$, with all other settings unchanged. We train both RL with pointwise reward and our method (RL with pairwise reward) on the same 5 OOD scenarios (in-place spinning, left turn, right turn, vehicle rollover, and doubled speed) for 200 steps each. As shown in Fig.\ref{fig:pointwise_pairwise_comparison}, our pairwise reward mechanism produces vehicles with notably better geometry, texture fidelity, and lighting plausibility, especially in the highlighted regions. This corroborates our analysis in Sec.3.2: pointwise scoring is susceptible to reward hacking\cite{Pref-GRPO&UniGenBench}, where negligible absolute score differences are amplified into misleading gradients. In contrast, our pairwise win-rate mechanism aggregates $\binom{N}{2}$ robust comparisons into a discriminative per-candidate reward, providing reliable optimization directions even among similar-quality candidates.

\end{document}